\documentclass[letterpaper, 10 pt, conference]{ieee/ieeeconf}  

\IEEEoverridecommandlockouts                              

\usepackage{booktabs}
\usepackage{fancyhdr}
\usepackage{graphics}
\usepackage{graphicx}
\usepackage{epsfig}
\usepackage{caption}
\usepackage{subcaption}
\usepackage{wrapfig}
\usepackage{float}
\usepackage{cite}
\usepackage{times}
\usepackage{amsmath}
\usepackage{amssymb}
\usepackage{amsfonts}
\usepackage{mathtools}
\usepackage[format=plain,
            labelfont=it,
            labelsep=period]{caption}
\usepackage{url}
\usepackage{algorithm}
\usepackage{algpseudocode}

\usepackage{color}
\usepackage{xcolor}
\definecolor{citecolor}{HTML}{0071bc}
\usepackage[pagebackref=true,breaklinks=true,colorlinks=true,bookmarks=false,linkcolor=citecolor]{hyperref}
\usepackage{cleveref}

\title{\LARGE \bf
Generalization in Reinforcement Learning by Soft Data Augmentation
}

\author{Nicklas Hansen$^{*\dagger}$, Xiaolong Wang$^{*}$%
\thanks{$^{*}$University of California, San Diego, CA, USA}%
\thanks{$^{\dagger}$Technical University of Denmark, Denmark}%
}

\begin{document}

\maketitle
\thispagestyle{empty}
\pagestyle{empty}

\begin{abstract}
Extensive efforts have been made to improve the generalization ability of Reinforcement Learning (RL) methods via domain randomization and data augmentation. However, as more factors of variation are introduced during training, optimization becomes increasingly challenging, and empirically may result in lower sample efficiency and unstable training. Instead of learning policies directly from augmented data, we propose SOft Data Augmentation (SODA), a method that decouples augmentation from policy learning. Specifically, SODA imposes a soft constraint on the encoder that aims to maximize the mutual information between latent representations of augmented and non-augmented data, while the RL optimization process uses strictly non-augmented data. Empirical evaluations are performed on diverse tasks from DeepMind Control suite as well as a robotic manipulation task, and we find SODA to significantly advance sample efficiency, generalization, and stability in training over state-of-the-art vision-based RL methods.\footnote{Webpage: \url{https://nicklashansen.github.io/SODA/}}
\end{abstract}

\section{INTRODUCTION}
\label{sec:introduction}
Reinforcement Learning (RL) from visual observations has achieved tremendous success in many areas, including game-playing~\cite{mnih2013playing}, robotic manipulation~\cite{levine2016end,nair2018visual,Kalashnikov2018QTOptSD}, and navigation tasks~\cite{zhu2017target,WeiYang_ICLR2019}. While this combination of learning both representation and decision-making jointly with RL has led to great success, numerous studies have shown that RL agents fail to generalize to new environments~\cite{Gamrian2019TransferLF, cobbe2018quantifying, laskin2020reinforcement, song2019observational, Cobbe2019LeveragingPG}, which is only exacerbated by the high-dimensional nature of images in practical applications.

To improve generalization in RL, domain randomization~\cite{pinto2017asymmetric,Tobin_2017,Peng_2018,yang2019single} has been widely applied to learn robust representations invariant to visual changes. The assumption made in these approaches is that the distribution of randomized environments sufficiently cover factors of variation encountered at test-time. However, as more factors of variation are introduced during training, RL optimization becomes increasingly challenging, and empirically may lead to lower sample efficiency and unstable training. If we constrain the randomization to be small, the distribution is unlikely to cover variations in the test environment. Finding the right balance between sample efficiency and generalization therefore requires extensive engineering efforts.

Similarly, recent studies also show that by using the right kinds of data augmentation, both sample efficiency and generalization of RL policies can be improved substantially~\cite{Lee2019ASR,laskin2020reinforcement,Raileanu2020AutomaticDA}. Determining which data augmentations to use (and to which degree) is however found to be task-dependent, and a naïve application of data augmentation may lead to unstable training and poor performance \cite{laskin2020reinforcement, Raileanu2020AutomaticDA}. It is therefore natural to ask whether we can obtain the benefits of data augmentation while interfering minimally with the RL optimization process.
\begin{figure}
    \centering
    \begin{subfigure}[b]{0.44\textwidth}
        \centering
        \includegraphics[width=\textwidth]{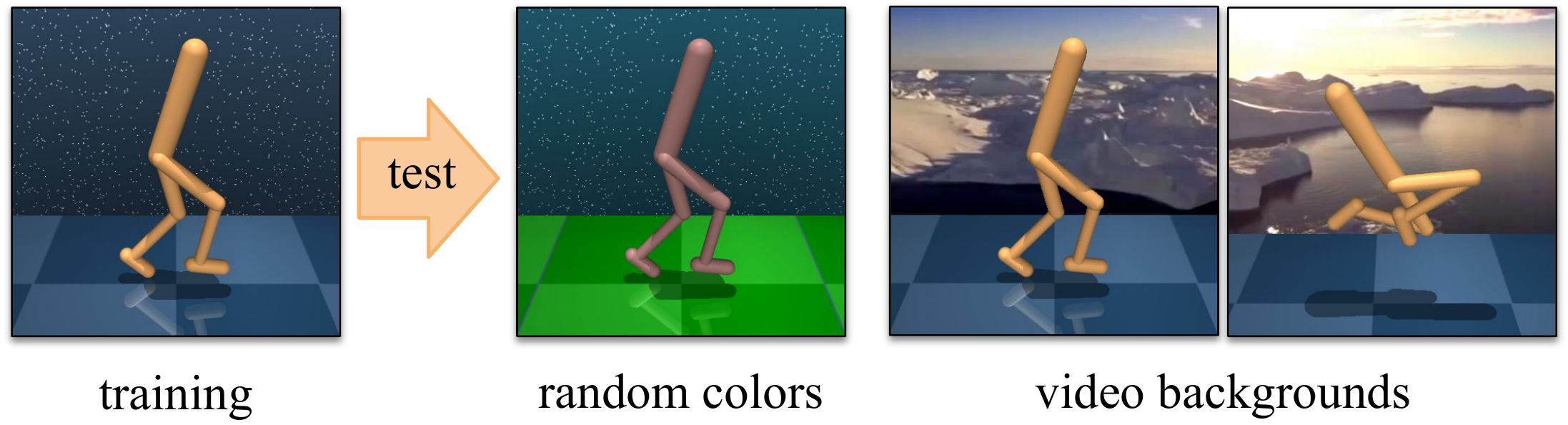}
        \caption{Environments for DeepMind Control tasks. We consider 5 challenging continuous control tasks from this benchmark.}
        \label{fig:samples-dmc}
    \end{subfigure}
    \begin{subfigure}[b]{0.44\textwidth}
        \centering
        \vspace{1ex}
        \includegraphics[width=\textwidth]{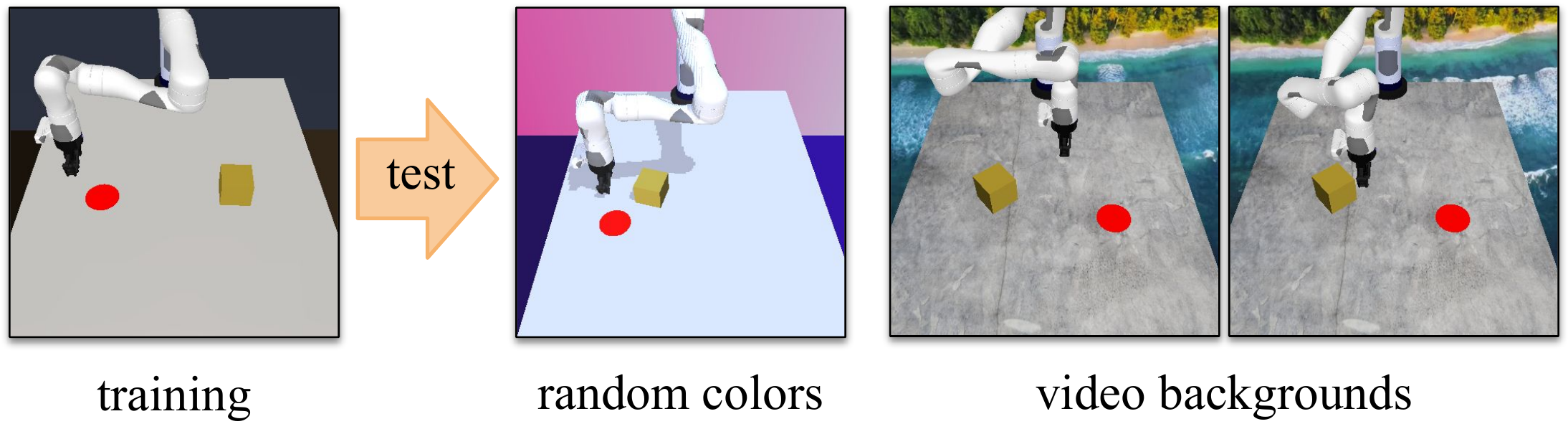}
        \caption{Environments for robotic manipulation. The task is to push the yellow cube to the location of the red disc.}
        \label{fig:samples-gen3}
    \end{subfigure}%
    \caption{\textbf{Generalization in RL.} Agents are trained in a fixed environment (denoted the \emph{training} environment) and we measure generalization to unseen environments with (i) \emph{random colors} and (ii) \emph{video backgrounds}. To simulate real-world deployment, we additionally randomize camera, lighting, and texture during evaluation in the robotic manipulation task. Additional samples are shown in appendix \ref{sec:appendix-dmc-gb}. }
    \vspace{-0.2in}
    \label{fig:teaser}
\end{figure}

In this paper, we propose \textbf{SO}ft \textbf{D}ata \textbf{A}ugmentation (SODA), a method that stabilizes training by decoupling data augmentation from policy learning. While previous work attempts to learn policies directly from augmented data, SODA uses strictly \emph{non-augmented} data for policy learning, and instead performs auxiliary representation learning using \emph{augmented} data. Through its auxiliary task, SODA learns to generalize by maximizing the mutual information between latent representations of augmented and non-augmented data. The proposed auxiliary task shares a learned encoder with the policy, and SODA therefore imposes a soft constraint on the learned representation. As the policy is learned only from non-augmented data, the difficulty of RL optimization is greatly reduced, while SODA still benefits substantially from data augmentation through representation learning.

We perform visual representation learning by projecting augmented observations into a compact latent space using the shared encoder and a learned projection, and similarly projecting non-augmented observations using a moving average of the learned network. The SODA objective is then to learn a mapping from latent features of the augmented observation to those of the original observation. With strong and varied data augmentation, SODA learns to ignore factors of variation that are irrelevant to the RL task, which greatly reduces observational overfitting \cite{song2019observational}. Our visual representation learning task is a self-supervised learning task. Our approach is related to recent work on joint learning with self-supervision and RL~\cite{yarats2019improving,srinivas2020curl,stooke2020atc}, where two tasks are trained jointly on the same augmented observations. However, our method is fundamentally different. We decouple the training data flow by using non-augmented data for RL and using augmented data only for representation learning. Besides, instead of learning invariance by contrasting two augmented instances of the same image to a batch of negative samples~\cite{srinivas2020curl,chen2020simple,He2020MomentumCF}, SODA learns to map augmented images to their non-augmented counterparts in latent space, without the need for negative samples. Because we strictly use data augmentation for representation learning and instead impose a soft constraint on the shared encoder through joint training, we call it \emph{soft} data augmentation.

Empirical evaluations are performed on the novel \emph{DMControl Generalization Benchmark} (DMControl-GB) based on continuous control tasks from DeepMind Control suite (DMControl)~\cite{deepmindcontrolsuite2018}, as well as a robotic manipulation task. We train in a fixed environment and evaluate generalization to environments that are unseen during training, as shown in Figure~\ref{fig:teaser}. Our method improves both sample efficiency and generalization over state-of-the-art vision-based RL methods in 9 out of 10 environments from DMControl-GB and all 3 types of test environments in robotic manipulation.

We highlight our main contributions as follows: 
\begin{itemize}
    \item{We present SODA, a stable and efficient representation learning method for vision-based RL.}
    \item{We propose the \emph{DMControl Generalization Benchmark}, a new benchmark for generalization in vision-based RL.}
    \item{We show that SODA outperforms prior state-of-the-art methods significantly in test-time generalization to unseen and visually diverse environments.}
\end{itemize}

\section{RELATED WORK}
\label{sec:related-work}

\textbf{Learning Visual Invariance with Self-Supervision.} Self-supervision for visual representation learning has proven highly successful in computer vision~\cite{wang2015unsupervised,doersch2015unsupervised,pathak2016context,noroozi2016unsupervised,zhang2016colorful}. Recently, contrastive learning for self-supervised pre-training has achieved very compelling results for many downstream visual recognition tasks~\cite{wu2018unsupervised,oord2018representation,He2020MomentumCF,tian2019contrastive,misra2020pirl,chen2020simple}. For example, Chen et al.~\cite{chen2020simple} perform an extensive study on different augmentations (e.g. random cropping, color distortion, rotation) and show that learned representations invariant to these transformations can make good initializations for downstream tasks. Instead of learning with a contrastive objective, Grill et al.~\cite{Grill2020BootstrapYO} show that invariance can also be learned via a prediction error without using any negative samples. Our method also learns invariant feature representations, but rather than aligning two different augmented views of the same instance as in previous work, we instead learn to align the representation of augmented and non-augmented views. As in Grill et al., SODA does not require negative samples.

\textbf{Self-Supervised Visual Learning for RL.} Inspired by the success of self-supervised visual representation learning, researchers have proposed to perform joint learning with self-supervision and RL to improve sample efficiency and performance~\cite{yarats2019improving,schwarzer2020data,srinivas2020curl}. For example, Yarats et al. ~\cite{yarats2019improving} show that jointly training an auto-encoder together with RL improves the sample efficiency of model-free RL. Srinivas et al.~\cite{srinivas2020curl} further extend this idea by performing auxiliary contrastive learning together with RL, and nearly match state-based RL in sample efficiency. While the proposed contrastive learning task uses data augmentation in training, the same augmented data is used for RL as well. In contrast to these approaches, we separate the training data into two data streams and strictly use the non-augmented data for RL, which helps stabilize training, decrease sensitivity to the choice of data augmentation, and allows us to benefit from stronger data augmentation in the self-supervised learning. Recent work also attempts to close the generalization gap by adapting a learned policy using observations collected from a target environment~\cite{Carr2019DomainAF, Chen2020AdversarialFT, Julian2020NeverSL}. Instead, SODA learns a \emph{single} policy that generalizes \emph{without} data from test environments.

\textbf{Data Augmentation and Randomization.} Our work is inspired by research on generalization by domain randomization~\cite{Tobin_2017, pinto2017asymmetric, Peng_2018, Ramos_2019, yang2019single}. For example, Tobin et al.~\cite{Tobin_2017} render scenes with random textures for object localization in simulation, and show that the learned networks transfer to the real world. Similar generalization across environments have been shown in recent work on RL with data augmentation~\cite{Lee2019ASR, laskin2020reinforcement,kostrikov2020image, Raileanu2020AutomaticDA}. In these works, policies are trained to be invariant to certain visual perturbations by augmenting observations. However, it is unpredictable and task-dependent how much randomization can be applied. Instead of training RL policies directly on augmented observations, we propose \emph{soft} data augmentation that decouples training into two objectives with separate data streams.

\begin{figure*}[t!]
    \centering
    \includegraphics[scale=0.625]{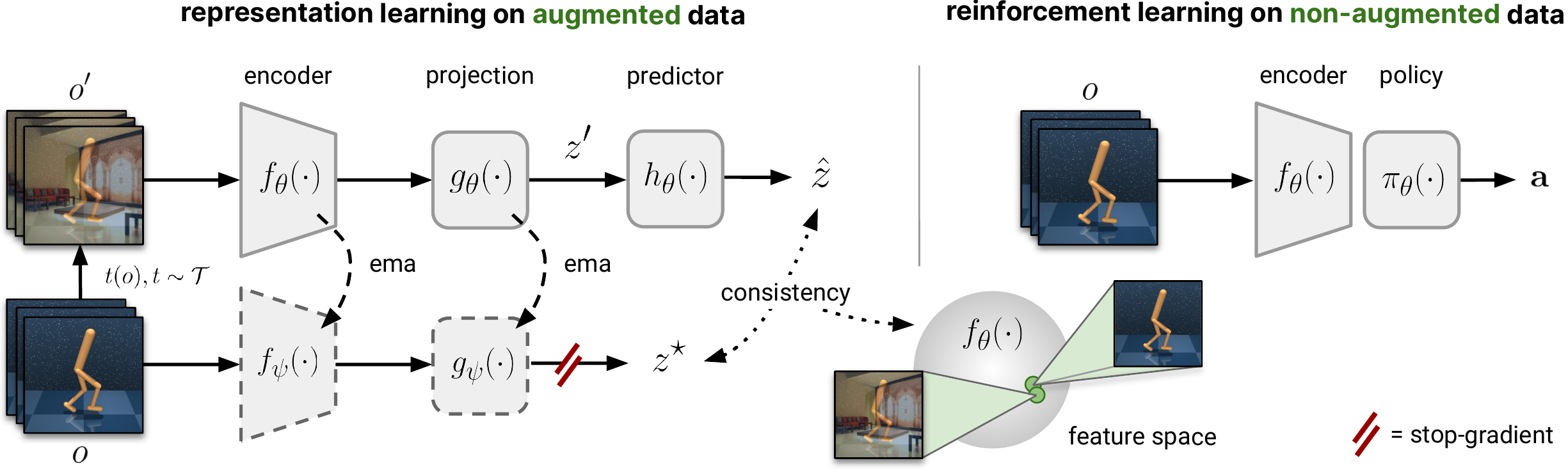}
    \caption{\textbf{SODA architecture.} \textit{Left:} an observation $o$ is augmented to produce a view $o'$, which is then encoded and projected into $z'=g_{\theta}(f_{\theta}(o'))$. Likewise, $o$ is encoded by $f_{\psi}$ and projected by $g_{\psi}$ to produce features $z^{\star}$. The SODA objective is then to predict $z^{\star}$ from $z'$ by $h_{\theta}$ formulated as a consistency loss. \textit{Right:} Reinforcement Learning in SODA. The RL task remains unchanged and is trained directly on the non-augmented observations $o$. \emph{ema} denotes an exponential moving average.}
    \label{fig:soda-architecture}
\end{figure*}

\section{METHOD}
\label{sec:method}
We propose \textbf{SO}ft \textbf{D}ata \textbf{A}ugmentation (SODA), a novel approach to data augmentation in RL that exhibits significant improvements in sample efficiency and stability over prior work. SODA is a general framework for data augmentation that can be implemented on top of any standard RL algorithm. It aims to learn representations that effectively encode information shared between an observation and its augmented counterpart, while interfering minimally with the RL objective. In the following, we present the individual components of SODA in detail.

\subsection{Architectural Overview}
\label{sec:architectural-overview}
SODA is implemented as a self-supervised auxiliary task that shares a common encoder $f$ with an RL policy. For a given policy network parameterized by a collection of parameters $\theta$, we split the network and corresponding parameters sequentially into an encoder $f_{\theta}$ and a policy $\pi_{\theta}$ such that $\mathbf{a} = \pi_{\theta}(f_{\theta}(o))$ outputs a distribution over actions (and any other algorithm-specific values) for an input observation $o$.
SODA then consists of the following three components: the shared encoder $f_{\theta}$, a projection $g_{\theta}$, and a prediction head $h_{\theta}$. We additionally consider an architecturally identical encoder $f_{\psi}$ and projection $g_{\psi}$ where $\psi$ denotes a collection of parameters separate from $\theta$.
We denote $f_{\theta}$ as the \emph{online} encoder and $f_{\psi}$ as the \emph{target} encoder, and similarly for projections $g_{\theta}, g_{\psi}$. We then define the parameters $\psi$ as an exponential moving average (EMA) of $\theta$, and update $\psi$ with every iteration of SODA using the update rule,
\begin{equation}
    \psi_{n+1} \longleftarrow (1-\tau) \psi_{n} + \tau \theta_{n}
\end{equation}
for an iteration step $n$ and a momentum coefficient $\tau \in(0,1]$, such that only parameters $\theta$ are updated by gradient descent~\cite{Lillicrap2016ContinuousCW, He2020MomentumCF, Grill2020BootstrapYO}. Figure~\ref{fig:soda-architecture} provides an overview of the architecture, and the SODA task is detailed in the following.

\subsection{Representation Learning by SODA}
\label{sec:soda-objective}
Given an observation $o\in\mathbb{R}^{C \times H \times W}$, we sample a data augmentation $t \sim \mathcal{T}$ and apply it to produce an augmented observation $o' \triangleq t(o)$. As such, $o$ and $o'$ can be considered different \emph{views} of the same observation, where $o$ is the original observation and $o'$ is corrupted by some noise source. If we assume $f : \mathbb{R}^{C \times H \times W} \rightarrow \mathbb{R}^{C' \times H' \times W'}$ such that $H' \times W'$ is a feature map down-sampled from $H \times W$, then both projections $g_{\theta}, g_{\psi}$ are learned mappings $g : \mathbb{R}^{C' \times H' \times W'} \rightarrow \mathbb{R}^{K}$ where $K \ll C' \times H' \times W'$. Given a feature vector $z' \triangleq g_{\theta}(f_{\theta}(o'))$, the proposed SODA task is then to learn a mapping $h_{\theta} : \mathbb{R}^{K} \rightarrow \mathbb{R}^{K}$ that predicts the target projection $z^{\star} \triangleq g_{\psi}(f_{\psi}(o))$ of the non-augmented observation $o$, as shown in Figure~\ref{fig:soda-architecture} (left). We optimize projection $g_{\theta}$, predictor $h_{\theta}$, and shared encoder $f_{\theta}$ jointly together with the RL task(s), and employ a consistency loss
\begin{equation}
    \label{eq:consistency-loss}
    \mathcal{L}_{SODA} \left(\hat{z}, z^{\star}; \theta \right) =
    \mathop{\mathbb{E}_{t\sim\mathcal{T}}}\left[ \left\| \hat{z}_{\circ} - z^{\star}_{\circ} \right\|^{2}_{2} \right]
\end{equation}
where $\hat{z} \triangleq h_{\theta}(z')$, and $\hat{z}_{\circ} \triangleq \hat{z} / \| \hat{z} \|_{2},~z^{\star}_{\circ} \triangleq z^{\star} / \| z^{\star} \|_{2}$ are $\ell_{2}$-normalizations of $\hat{z}$ and $z^{\star}$, respectively. We use a predictor $h_{\theta}$ rather than mapping $z'$ directly to $z^{\star}$ as we find it to improve expressiveness in the learned representations, which is consistent with prior work \cite{chen2020simple, Grill2020BootstrapYO, schwarzer2020data}.

The policy $\pi_{\theta}$ (including any algorithm-specific task heads) uses $f_{\theta}$ to extract features and is optimized with no modifications to architecture nor inputs, i.e. we optimize the RL objective $\mathcal{L}_{RL}$ directly on non-augmented observations $o$ and update $\pi_{\theta}, f_{\theta}$ by gradient descent, as shown in Figure~\ref{fig:soda-architecture} (right). During training, we continuously alternate between optimizing $\mathcal{L}_{RL}$ and $\mathcal{L}_{SODA}$. At test-time, we only use the encoder $f_{\theta}$ and policy $\pi_{\theta}$. The training procedure is summarized in Algorithm \ref{alg:soda}, and we further motivate the algorithmic design in \cref{sec:mutual-information}.
\begin{algorithm}
\caption{Soft Data Augmentation (SODA)}
\label{alg:soda}
\begin{algorithmic}[1]
\algnotext{EndFor}
\Statex $\theta, \psi$: randomly initialized network parameters
\Statex $\omega$: RL updates per iteration
\Statex $\tau$: momentum coefficient
\For{every iteration}
\For{$\textnormal{update}=1,2,...,\omega$}
\State Sample batch of transitions $\nu \sim \mathcal{B}$
\State Optimize $\mathcal{L}_{RL} \left(\nu\right)$ wrt $\theta$
\EndFor
\State Sample batch of observations $o \sim \mathcal{B}$
\State Augment observations $o' = t(o),~t \sim \mathcal{T}$
\State Compute online predictions $\hat{z} = h_\theta(g_{\theta}(f_{\theta}(o')))$ 
\State Compute target projections $z^{\star} = g_{\psi}(f_{\psi}(o))$
\State Optimize $\mathcal{L}_{SODA} \left(\hat{z}, z^{\star}\right)$ wrt $\theta$
\State Update $\psi \leftarrow (1-\tau) \psi + \tau \theta$
\EndFor
\end{algorithmic}
\end{algorithm}
\subsection{Data Augmentation as a Mutual Information Problem}
\label{sec:mutual-information}
SODA reformulates the problem of generalization as a representational consistency learning problem: the encoder $f$ should learn to map different views of the same underlying state to similar feature vectors. This encourages the encoder to learn features that are shared across views, e.g. physical quantities and object interactions, and discard information that yields no predictive power such as background, lighting, and high-frequency noise. Given an observation $o$ and a data augmentation $t$, we seek to encode $o$ and $o'=t(o)$ into compact feature vectors $z', z^{\star} \in \mathbb{R}^{K}$, respectively, by learned mappings $f_{\theta}, g_{\theta}$ and $f_{\psi}, g_{\psi}$ such that the mutual information $I$ between $o$ and $o'$ is maximally preserved. The mutual information between $z^{\star}$ and $z'$ is then given by
\begin{equation}
    I(z^{\star}; z') = \sum_{z'} \sum_{z^{\star}} p(z^{\star}, z') \log \frac{p(z^{\star} | z')}{p(z^{\star})}
\end{equation}
\begin{figure}
    \centering
    \includegraphics[width=0.3\textwidth]{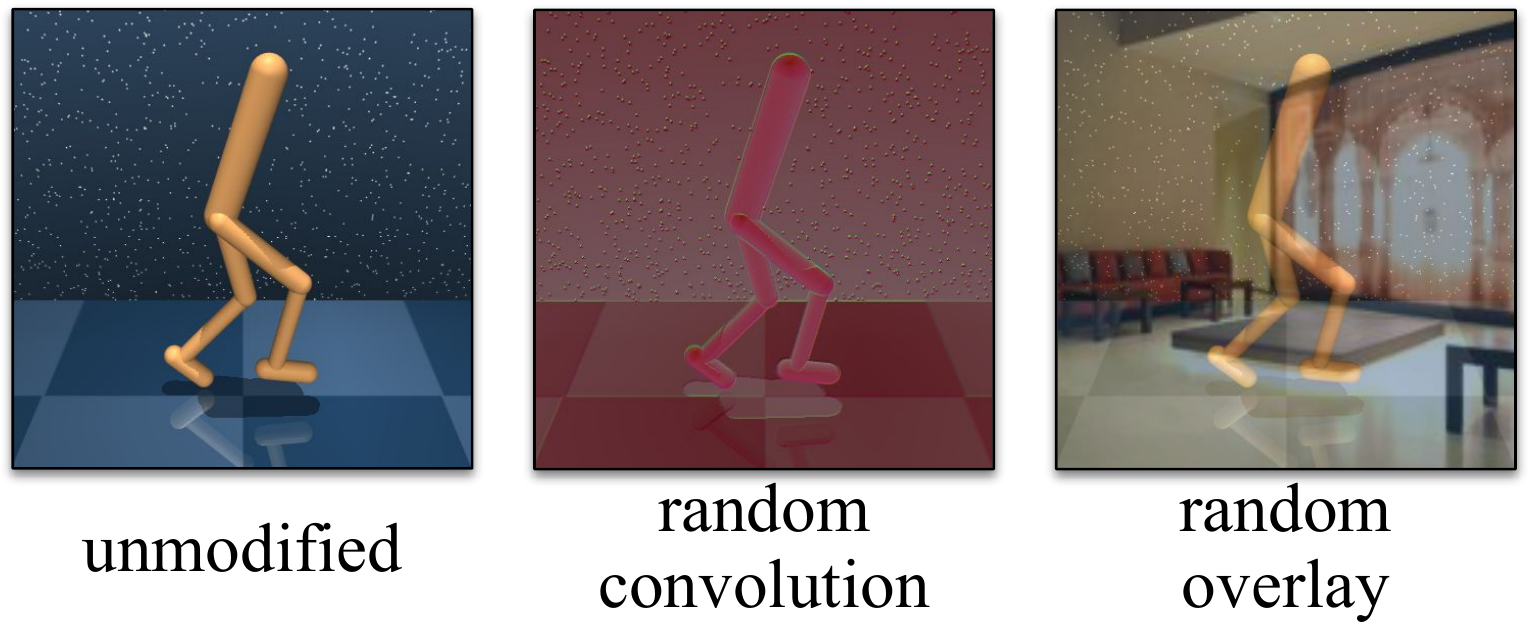}
    \caption{\textbf{Data augmentation.} We consider the following two data augmentations: \emph{random convolution} (as proposed by \cite{Lee2019ASR, laskin2020reinforcement}) and \emph{random overlay} (novel). See appendix~\ref{sec:appendix-data-augmentations} for additional data augmentation samples.}
    \label{fig:dmc-aug-samples}
    \vspace{-0.1in}
\end{figure}
and is naturally bounded by the mutual information between $o$ and $o'$, i.e. $I(z^{\star}; z') \le I(o; o')$. As $I(z^{\star}; z')$ is impractical to optimize directly, we approximate $I$ as a simple consistency loss defined in (\ref{eq:consistency-loss}) by the following intuition. If we assume $f_{\psi}, g_{\psi}$ to maximally preserve information in $o$, then $I(z^{\star}; z') \propto I(o; o')$, and minimizing $\mathcal{L}_{SODA}$ thus maximizes $I(z^{\star}; z' | o')$. In other words, by learning a mapping from $z'$ to $z^{\star}$, we maximally preserve information in $o$ by learning to discard noise added by the data augmentation $t(\cdot)$. With strong and varied data augmentation, $f_{\theta}$ learns to ignore factors of variation that are irrelevant to the RL task, and reduces observational overfitting \cite{song2019observational, packer2018assessing}. While the assumption of maximally preserved information may not hold in practice, we find Equation \ref{eq:consistency-loss} to be a good enough approximation for expressive representation learning. Although the SODA objective does not explicitly prevent trivial solutions, e.g. $f_{\theta}(\cdot) = \mathbf{0}$, such behavior is implicitly discouraged, and the reason for that is two-fold: (i) trivial equilibria in $\mathcal{L}_{SODA}$ are shown to be unstable due to updates in the targets $f_{\psi}, g_{\psi}$ \cite{Grill2020BootstrapYO}; and (ii) $f_{\theta}$ is jointly optimized between SODA and the RL objective(s), and trivial equilibria are in direct conflict with the RL objective since they entail that $f_{\theta}$ then would preserve no information in $o$. While the predictor $h_{\theta}$ is not strictly needed, we empirically find it to improve the sample efficiency of our method, which we conjecture is due to non-stationarity of the target projection $z^{\star}$.
\begin{figure*}[t]
    \centering
    \includegraphics[width=\linewidth]{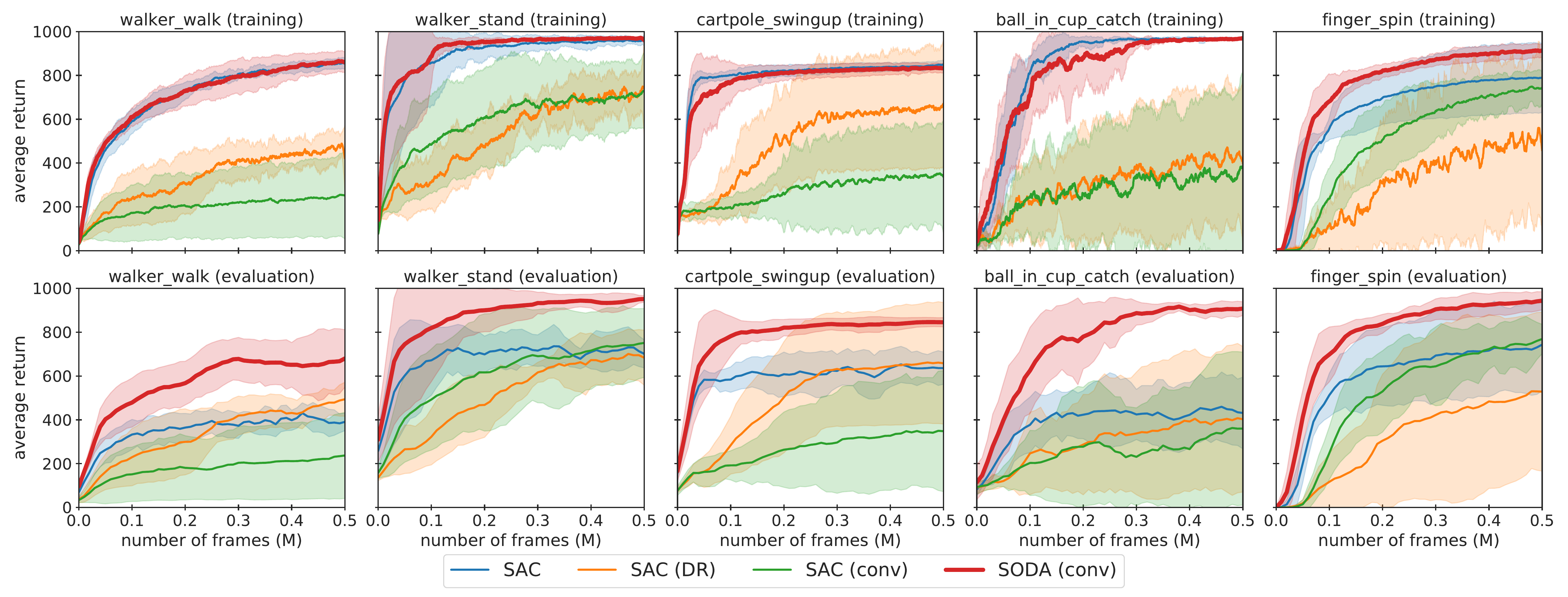}
    \caption{\textbf{Random convolution.} \textit{Top:} average return on the training environment during training. \textit{Bottom:} periodic evaluation of generalization ability measured by average return on the \emph{random color} environment. SODA exhibits sample efficiency and convergence similar to SAC but improves generalization significantly. Average of 5 runs, shaded area is std. deviation.}
    \label{fig:dmc-curves-conv}
    \vspace{-0.1in}
\end{figure*}
\section{EXPERIMENTS}
\label{sec:experiments}
We evaluate SODA on 5 tasks from DeepMind control suite (DMControl) \cite{deepmindcontrolsuite2018}, as well as in robotic manipulation. DMControl is comprised of diverse and challenging continuous control tasks and is a widely used benchmark for vision-based RL \cite{yarats2019improving, srinivas2020curl, laskin2020reinforcement, kostrikov2020image, hansen2020deployment}. To quantify the generalization ability of SODA, we propose \emph{DMControl Generalization Benchmark} (DMControl-GB)\footnote{SODA and benchmark is open-sourced at: \url{https://github.com/nicklashansen/dmcontrol-generalization-benchmark}}, a new benchmark for generalization in vision-based RL, based on DMControl. Agents are trained in a fixed environment (denoted the \emph{training} environment), and we measure generalization to two distinct test distributions: (1) environments with randomized colors; and (2) environments with natural videos as background, first proposed by \cite{hansen2020deployment}. These test distributions correspond to the \texttt{color\_hard} and \texttt{video\_easy} benchmarks from DMControl-GB; we provide results for the full DMControl-GB benchmark in appendix \ref{sec:appendix-dmc-gb}.

While DMControl-GB provides a good platform for benchmarking algorithms, we are ultimately interested in developing algorithms that solve real-world problems with vision-based RL. To better emulate real-world deployment scenarios, we also consider a robotic manipulation task on a robotic arm (in simulation). As in DMControl-GB, agents are trained in a fixed environment and evaluated on environments with randomized colors and video backgrounds, and we additionally perform random perturbations of camera, lighting, and texture to simulate real-world conditions during testing. Samples from DMControl-GB and robotic manipulation are shown in Figure~\ref{fig:teaser}, and additional samples are shown in appendix~\ref{sec:appendix-dmc-gb}.

\subsection{Implementation Details}
SODA is a general framework that can be implemented on top of any standard RL algorithm using parameterized policies. In this work we implement SODA using Soft Actor-Critic (SAC) \cite{sacapps} as the base algorithm. We adopt network architecture and hyperparameters from \cite{hansen2020deployment}, which implements $f_{\theta}$ as 11 convolutional layers and $\pi_{\theta}$ as a multi-layer perceptron (MLP). Both $g_{\theta}, h_{\theta}$ are implemented as MLPs with batch normalization \cite{Ioffe2015BatchNA} and we use a batch size of 256 for the SODA task. Projections are of dimension $K=100$, and the target components $f_{\psi}, g_{\psi}$ are updated with a momentum coefficient $\tau = 0.005$. $\mathcal{L}_{RL}$ and $\mathcal{L}_{SODA}$ are optimized using Adam \cite{Kingma2015AdamAM}, and in practice we find it sufficient to only make a SODA update after every second RL update, i.e. $\omega=2$. In all tasks considered, observations are stacks of $k$ consecutive frames ($k=3$ for DMControl-GB and $k=1$ for robotic manipulation) of size $100\times100$ and with no access to state information. See appendix~\ref{sec:appendix-hyperparams} for further implementation details and a list of hyper-parameters.

\subsection{Data Augmentation}
\label{sec:data-augmentation}
While SODA makes no assumptions about the data augmentation used, it is useful to distinguish between \emph{weak} augmentations that may improve sample efficiency but contribute minimally to generalization, and \emph{strong} augmentations that improve generalization at the cost of sample efficiency \cite{Tian2020WhatMF}. An example of the former is random cropping \cite{kostrikov2020image, laskin2020reinforcement}, whereas examples of the latter are domain randomization \cite{Tobin_2017, pinto2017asymmetric} and random convolution \cite{Lee2019ASR, laskin2020reinforcement}. Consistent with previous work \cite{srinivas2020curl, laskin2020reinforcement, hansen2020deployment}, we apply temporally consistent random cropping (down to $84\times84$) by default in SODA and all baselines, and we refer to the cropped images as non-augmented observations.

We additionally seek to stabilize training with \emph{strong} augmentations for generalization, and consider the following two data augmentations: \emph{random convolution}, where a randomly initialized convolutional layer is applied; and a novel \emph{random overlay} that linearly interpolates between an observation $o$ and an image $\varepsilon$ to produce an augmented view
\begin{equation}
    \label{eq:random-overlay}
    t_{overlay}(o) = (1-\alpha)o + \alpha \varepsilon,~\varepsilon\sim\mathcal{D}
\end{equation}
where $\alpha \in[0,1)$ is the interpolation coefficient and $\mathcal{D}$ is an unrelated dataset. In practice, we use $\alpha=0.5$ and sample images from Places \cite{zhou2017places}, a dataset containing 1.8M diverse scenes. Samples of both data augmentations are shown in Figure~\ref{fig:dmc-aug-samples}, and additional samples are shown in Appendix~\ref{sec:appendix-data-augmentations}.

\begin{table*}[t]
\caption{\textbf{Generalization.} Average return of methods trained in a fixed environment and evaluated on: \textit{(left)} DMControl-GB with natural videos as background; and \textit{(right)} DMControl-GB with random colors. Mean and std. deviation of 5 runs.}
\label{tab:dmc-all}
\centering
\resizebox{0.925\textwidth}{!}{%
\begin{tabular}{lcccccccc||cccc}
& \multicolumn{8}{c}{\textbf{video backgrounds}} & \multicolumn{4}{c}{\textbf{random colors}} \\
\toprule
DMControl-GB                   & CURL & RAD & PAD  & SAC  & SAC    & SODA      & SAC          & SODA & CURL & RAD & PAD & SODA \\
(generalization)             & \cite{srinivas2020curl} & \cite{laskin2020reinforcement} & \cite{hansen2020deployment} & (DR) & (conv) & (conv)    & (overlay)    & (overlay) & \cite{srinivas2020curl} & \cite{laskin2020reinforcement} & \cite{hansen2020deployment} & (overlay) \\ \midrule
\texttt{walker,}        & $556$ & $606$ & $717$ & $520$ & $169$ & $635$ & $718$ & $\mathbf{768}$ & $445$ & $400$ & $468$ & $\mathbf{692}$ \vspace{-0.75ex} \\
\texttt{walk}           & $\scriptstyle{\pm133}$ & $\scriptstyle{\pm63}$ & $\scriptstyle{\pm79}$ & $\scriptstyle{\pm107}$ & $\scriptstyle{\pm124}$ & $\scriptstyle{\pm48}$ & $\scriptstyle{\pm47}$ & $\mathbf{\scriptstyle{\pm38}}$ & $\scriptstyle{\pm99}$ & $\scriptstyle{\pm61}$ & $\scriptstyle{\pm47}$ & $\mathbf{\scriptstyle{\pm68}}$ \vspace{0.75ex} \\
\texttt{walker,}        & $852$ & $745$ & $935$ & $839$ & $435$ & $903$ & $\mathbf{960}$ & $955$ & $662$ & $644$ & $797$ & $\mathbf{893}$ \vspace{-0.75ex} \\
\texttt{stand}          & $\scriptstyle{\pm75}$ & $\scriptstyle{\pm146}$ & $\scriptstyle{\pm20}$ & $\scriptstyle{\pm58}$ & $\scriptstyle{\pm100}$ & $\scriptstyle{\pm56}$ & $\mathbf{\scriptstyle{\pm2}}$ & $\scriptstyle{\pm13}$ & $\scriptstyle{\pm54}$ & $\scriptstyle{\pm88}$ & $\scriptstyle{\pm46}$ & $\mathbf{\scriptstyle{\pm12}}$ \vspace{0.75ex} \\
\texttt{cartpole,}      & $404$ & $373$ & $521$ & $524$ & $176$ & $474$ & $718$ & $\mathbf{758}$ & $454$ & $590$ & $630$ & $\mathbf{805}$ \vspace{-0.75ex} \\
\texttt{swingup}        & $\scriptstyle{\pm67}$ & $\scriptstyle{\pm72}$ & $\scriptstyle{\pm76}$ & $\scriptstyle{\pm184}$ & $\scriptstyle{\pm62}$ & $\scriptstyle{\pm143}$ & $\scriptstyle{\pm30}$ & $\mathbf{\scriptstyle{\pm62}}$ & $\scriptstyle{\pm110}$ & $\scriptstyle{\pm53}$ & $\scriptstyle{\pm63}$ & $\mathbf{\scriptstyle{\pm28}}$ \vspace{0.75ex} \\
\texttt{ball\_in\_cup,} & $316$ & $481$ & $436$ & $232$ & $249$ & $539$ & $713$ & $\mathbf{875}$ & $231$ & $541$ & $563$ & $\mathbf{949}$ \vspace{-0.75ex} \\
\texttt{catch}          & $\scriptstyle{\pm119}$ & $\scriptstyle{\pm26}$ & $\scriptstyle{\pm55}$ & $\scriptstyle{\pm135}$ & $\scriptstyle{\pm190}$ & $\scriptstyle{\pm111}$ & $\scriptstyle{\pm125}$ & $\mathbf{\scriptstyle{\pm56}}$ & $\scriptstyle{\pm92}$ & $\scriptstyle{\pm29}$ & $\scriptstyle{\pm50}$ & $\mathbf{\scriptstyle{\pm19}}$ \vspace{0.75ex} \\
\texttt{finger,}        & $502$ & $400$ & $691$ & $402$ & $355$ & $363$ & $607$ & $\mathbf{695}$ & $691$ & $667$ & $\mathbf{803}$ & $793$ \vspace{-0.75ex} \\
\texttt{spin}           & $\scriptstyle{\pm19}$ & $\scriptstyle{\pm64}$ & $\scriptstyle{\pm80}$ & $\scriptstyle{\pm208}$ & $\scriptstyle{\pm88}$ & $\scriptstyle{\pm185}$ & $\scriptstyle{\pm68}$ & $\mathbf{\scriptstyle{\pm97}}$ & $\scriptstyle{\pm12}$ & $\scriptstyle{\pm154}$ & $\mathbf{\scriptstyle{\pm72}}$ & $\scriptstyle{\pm128}$ \\\bottomrule
\end{tabular}
}
\vspace{-0.05in}
\end{table*}

\subsection{Baselines}
\label{sec:baselines}
We compare SODA to the following baselines: (i) a baseline SAC that is equivalent to RAD \cite{laskin2020reinforcement} as we apply random cropping to all baselines; (ii) SAC trained with domain randomization on the \emph{random colors} test distribution (denoted \emph{SAC (DR)}); (iii) SAC using random convolution as data augmentation (denoted \emph{SAC (conv)}; and (iv) SAC using random overlay as data augmentation (denoted \emph{SAC (overlay)}). Likewise, our method is denoted \emph{SODA (conv)} and \emph{SODA (overlay)}. Data augmentation baselines perform RL on augmented observations, whereas SODA does not; we ablate this decision in \cref{sec:stability}.

\subsection{Stability under Strong Data Augmentation}
\label{sec:stability}
Domain randomization and data augmentation is known to decrease sample efficiency and can destabilize the RL training. Figure~\ref{fig:dmc-curves-conv} shows training performance and generalization of baselines and SODA using random convolution. While SAC converges in training, its generalization remains poor. SODA improves generalization substantially without destabilizing the RL training, and exhibits a sample efficiency similar to the SAC baseline. Both SAC trained with domain randomization and \emph{SAC (conv)} either fail to solve the tasks or converge to sub-optimal policies.

\begin{figure}
    \centering
    \includegraphics[width=0.45\textwidth]{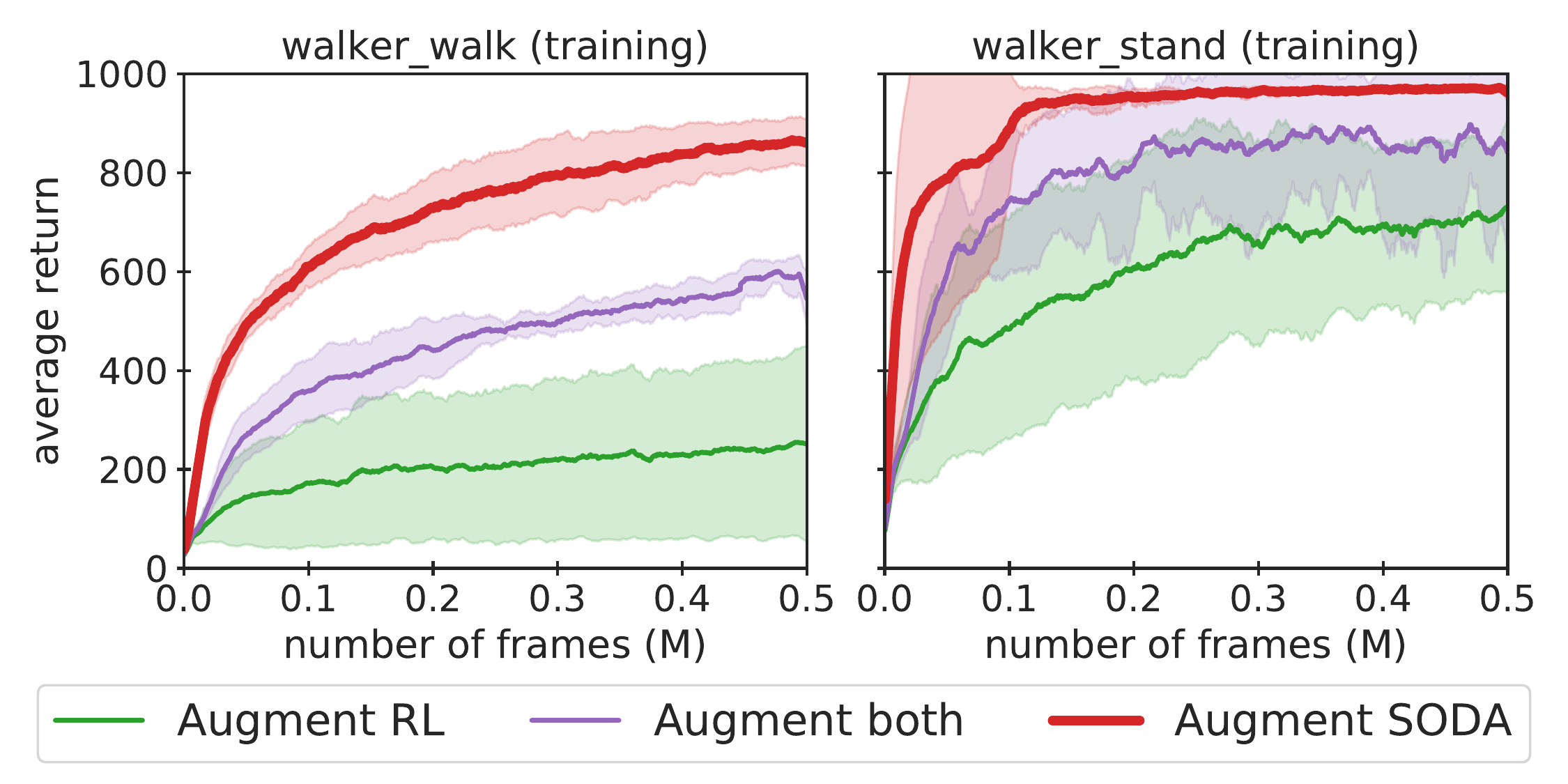}
    \caption{\textbf{Soft data augmentation.} Average return on the training environment for \texttt{walker\_walk} and \texttt{walker\_stand} tasks. \emph{Augment RL} corresponds to the SAC (conv) baseline, \emph{Augment both} applies random convolution in both SODA and RL, and \emph{Augment SODA} is the proposed formulation of SODA. Average of 5 runs, shaded area is std. deviation.}
    \label{fig:ablation}
    \vspace{-0.1in}
\end{figure}

To further investigate the effect of soft data augmentation, we consider a variant of SODA in which random convolution augmentation is applied in both representation learning and RL, and we summarize our findings in Figure~\ref{fig:ablation}. While using data augmentation in both objectives (\emph{Augment both}) improves significantly over only applying data augmentation in RL (\emph{Augment RL}), we find the proposed formulation of SODA (\emph{Augment SODA}) superior in both sample efficiency and end-performance.

\begin{figure*}[t]
    \centering
    \includegraphics[width=\linewidth]{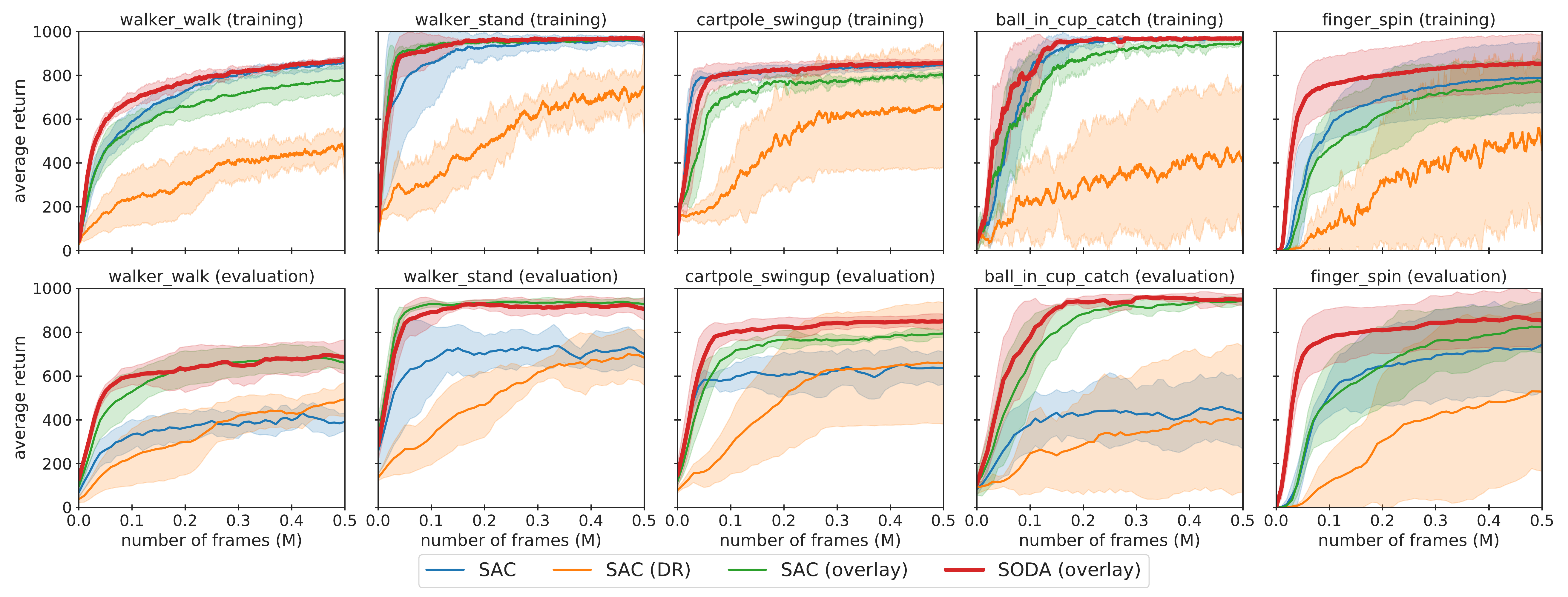}
    \caption{\textbf{Random overlay.} \textit{Top:} average return on the training environment during training. \textit{Bottom:} periodic evaluation of generalization ability measured by average return on the \emph{random color} environment. SODA offers better sample-efficiency than the novel \emph{SAC (overlay)} baseline and similar generalization to \emph{SODA (conv)} even though there is minimal visual similarity between random overlays and the random color environment. Average of 5 runs, shaded area is std. deviation.}
    \label{fig:dmc-curves-overlay}
\end{figure*}

\subsection{Generalization to Unseen Environments}
We evaluate the generalization ability of SODA on the challenging \emph{random colors} and \emph{video backgrounds} benchmarks from DMControl-GB (\texttt{color\_hard} and \texttt{video\_easy}, respectively), and additionally compare to a number of recent state-of-the-art methods: (i) CURL \cite{srinivas2020curl}, a contrastive representation learning method; (ii) RAD \cite{laskin2020reinforcement}, a study on data augmentation for RL that uses random cropping and is equivalent to the SAC baseline in \cref{sec:baselines}; and (iii) PAD \cite{hansen2020deployment}, a method for self-supervised policy adaptation. Results are shown in Table~\ref{tab:dmc-all}. We find SODA to outperform previous methods in \textbf{9} out of \textbf{10} instances, and by as much as \textbf{81\%} on \texttt{ball\_in\_cup\_catch} (video backgrounds). Further, SODA also outperforms the baselines from \cref{sec:baselines} in all tasks except \texttt{walker\_stand} on video backgrounds.

While SODA with random convolution data augmentation generalizes well to the random colors distribution, it generalizes comparably worse to videos. We conjecture that this is because random convolution captures less factors of variation than overlay (e.g. textures and local color changes), which is in line with the findings of \cite{cobbe2018quantifying, hansen2020deployment}. SODA (overlay) is found to generalize well to both test distributions even though there is minimal visual similarity between random overlays and the random color environment. This shows that SODA benefits from strong and varied data augmentation, and we leave it to future work to explore additional (soft) data augmentations. For completeness, training and generalization curves for the random overlay data augmentation are shown in Figure~\ref{fig:dmc-curves-overlay}, and we provide results and comparisons on two additional benchmarks, \texttt{color\_easy} and \texttt{video\_hard}, from DMControl-GB in appendix \ref{sec:appendix-dmc-gb}.

\begin{figure}
    \centering
    \includegraphics[width=0.475\textwidth]{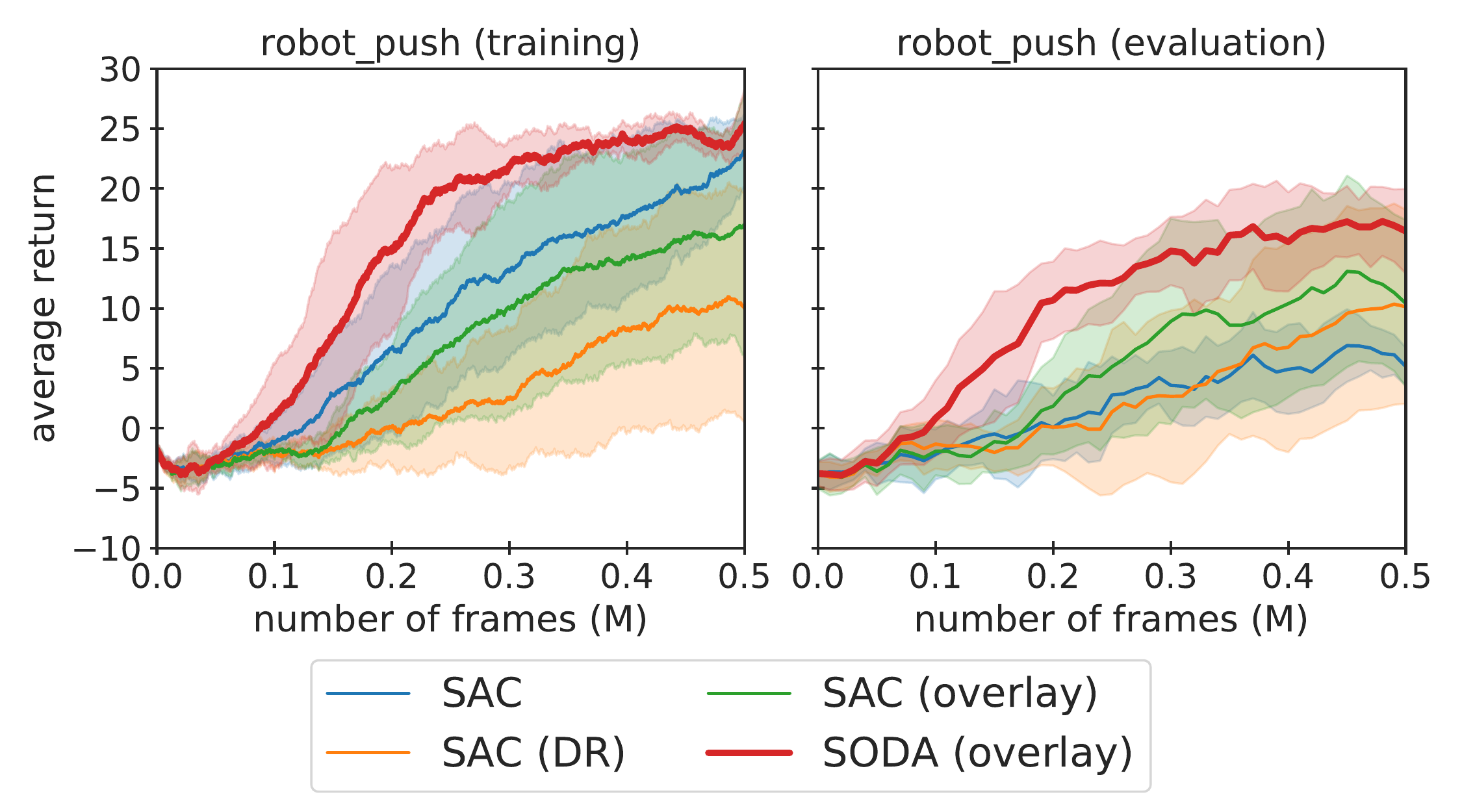}
    \caption{\textbf{Robotic manipulation.} \textit{Left:} average return on the training environment during training. \textit{Right:} periodic evaluation of generalization measured by average return on the \emph{random color} environment. SODA outperforms all baselines in both sample efficiency and generalization, and reduces variance. Average of 5 runs, shaded area is std. deviation.}
    \label{fig:robot-curves}
\end{figure}

\subsection{Robotic manipulation}
We additionally consider a robotic manipulation task in which the goal is for a robotic arm to push a yellow cube to the location of a red disc (as illustrated in Figure~\ref{fig:teaser}). Observations are $100\times100$ images and agents are trained using dense rewards. Episodes consist of 50 time steps, and at each time step there is a positive reward of $1$ for having the cube at the goal location (disc), and a small penalty proportional to the distance between the cube and goal. An average return of 25 therefore means that the cube is in goal position for more than half of all time steps.

\begin{table}
\caption{\textbf{Generalization in robotic manipulation.} Average return of SODA and baselines in the training and test environments. Mean and std. deviation of 5 runs.}
\label{tab:robot}
\centering
\resizebox{0.485\textwidth}{!}{%
\begin{tabular}{lcccccc}
\toprule
\texttt{robot,}         & SAC   & SAC  & SAC    & SODA      & SAC          & SODA \\
\texttt{push}           &       & (DR) & (conv) & (conv)    & (overlay)    & (overlay) \\ \midrule
training                & $14.5$ & $9.3$ & $0.2$ & $14.4$ & $15.8$ & $\mathbf{21.3}$ \vspace{-0.75ex} \\
                        & $\scriptstyle{\pm5.2}$ & $\scriptstyle{\pm7.5}$ & $\scriptstyle{\pm1.7}$ & $\scriptstyle{\pm4.6}$ & $\scriptstyle{\pm7.3}$ & $\mathbf{\scriptstyle{\pm5.0}}$ \vspace{0.75ex} \\
random                  & $7.2$ & $7.8$ & $-3.1$ & $1.4$ & $13.5$ & $\mathbf{18.0}$ \vspace{-0.75ex} \\
colors                  & $\scriptstyle{\pm3.4}$ & $\scriptstyle{\pm7.0}$ & $\scriptstyle{\pm1.4}$ & $\scriptstyle{\pm3.4}$ & $\scriptstyle{\pm4.4}$ & $\mathbf{\scriptstyle{\pm3.7}}$ \vspace{0.75ex} \\
video                   & $0.9$ & $7.0$ & $-4.9$ & $-4.5$ & $7.6$ & $\mathbf{13.9}$ \vspace{-0.75ex} \\
bg.                     & $\scriptstyle{\pm6.0}$ & $\scriptstyle{\pm7.4}$ & $\scriptstyle{\pm1.9}$ & $\scriptstyle{\pm1.1}$ & $\scriptstyle{\pm7.1}$ & $\mathbf{\scriptstyle{\pm2.6}}$ \\\bottomrule
\end{tabular}
}
\vspace{-0.05in}
\end{table}

As in DMControl-GB, we consider generalization to environments with (i) random colors; and (ii) video backgrounds. To simulate real-world deployment, we also randomize camera, lighting, and texture during testing. Training and generalization curves are shown in Figure~\ref{fig:robot-curves}, and results are summarized in Table~\ref{tab:robot}. We find SODA to outperform baselines in both sample efficiency and generalization to all considered test environments, which suggests that SODA may be a suitable method for reducing observational overfitting in real-world deployment of policies for robotic manipulation.

\section{CONCLUSION}
We show empirically that SODA improves stability, sample efficiency, and generalization significantly over previous methods and a set of strong baselines. We dissect the success of SODA and find that it benefits substantially from (i) not learning policies on augmented data; (ii) its representational consistency learning formulation; and (iii) strong and varied data augmentation for representation learning.
\newline
\section*{Acknowledgements}
This work was supported, in part, by grants from DARPA LwLL, NSF 1730158 CI-New: Cognitive Hardware and Software Ecosystem Community Infrastructure (CHASE-CI), NSF ACI-1541349 CC*DNI Pacific Research Platform, as well as gifts from Qualcomm and TuSimple.

\bibliography{references}
\bibliographystyle{ieee/IEEEtran}
\appendices
\section{DMCONTROL GENERALIZATION BENCHMARK}
\label{sec:appendix-dmc-gb}
\begin{figure}
    \centering
    \begin{subfigure}[b]{0.48\textwidth}
        \centering
        \includegraphics[width=\textwidth]{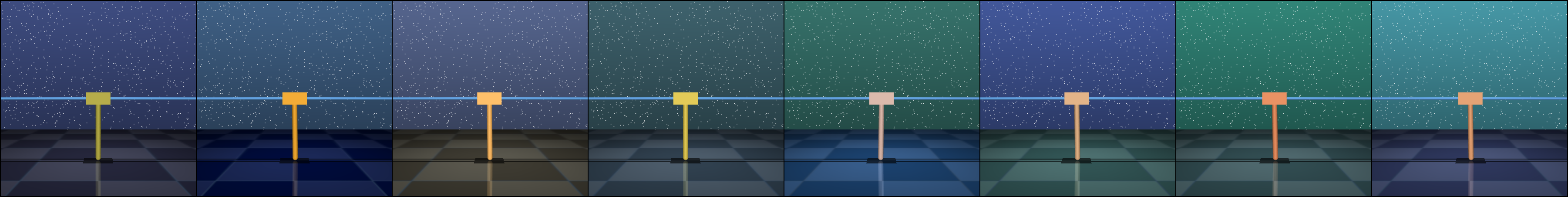}
        \caption{\texttt{color\_easy}}
        \label{fig:samples-dmc-gb-color-easy}
    \end{subfigure}
    \begin{subfigure}[b]{0.48\textwidth}
        \centering
        \vspace{2ex}
        \includegraphics[width=\textwidth]{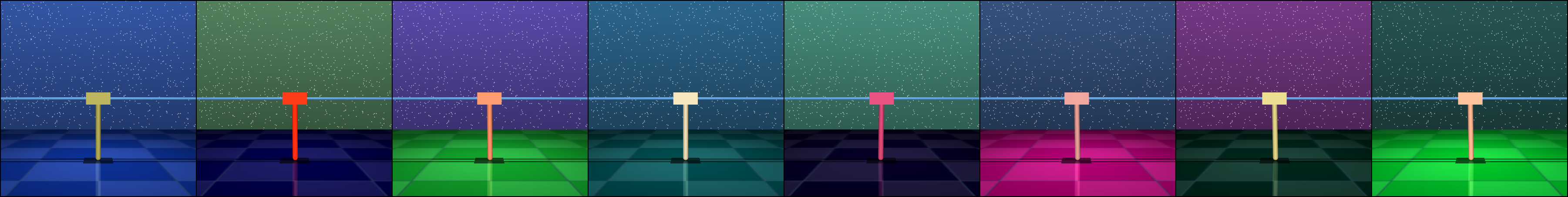}
        \caption{\texttt{color\_hard}}
        \label{fig:samples-dmc-gb-color-hard}
    \end{subfigure}
    \begin{subfigure}[b]{0.48\textwidth}
        \centering
        \vspace{2ex}
        \includegraphics[width=\textwidth]{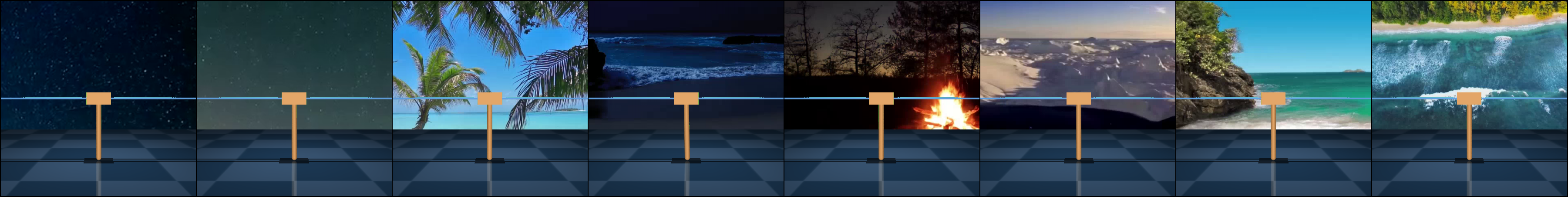}
        \caption{\texttt{video\_easy}}
        \label{fig:samples-dmc-gb-video-easy}
    \end{subfigure}
    \begin{subfigure}[b]{0.48\textwidth}
        \centering
        \vspace{2ex}
        \includegraphics[width=\textwidth]{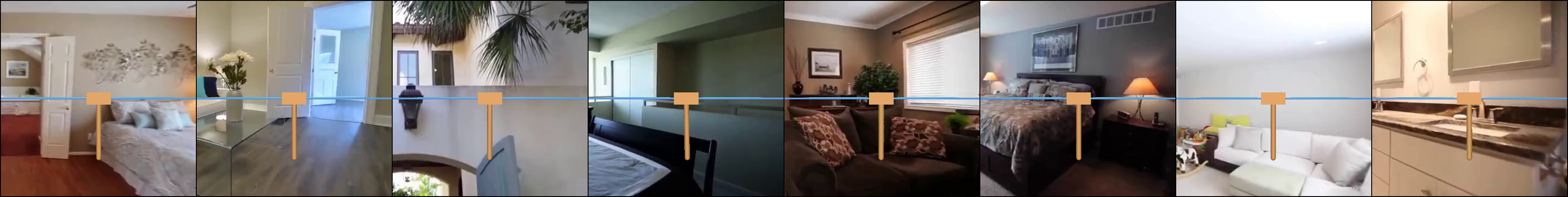}
        \caption{\texttt{video\_hard}}
        \label{fig:samples-dmc-gb-video-hard}
    \end{subfigure}%
    \caption{\textbf{Samples from DMControl-GB.} Agents are trained in a fixed environment (denoted the \emph{training} environment), and the generalization ability of agents is evaluated on the test distributions shown in (a-d). Environments (a-b) randomize the color of background, floor, and the agent itself, while (c-d) replaces the background with natural videos. In (c), only the skybox is replaced, whereas the entirety of the background is replaced in (d). Our main results, Figure \ref{fig:dmc-curves-conv}, Figure \ref{fig:dmc-curves-overlay}, and Table \ref{tab:dmc-all}, consider the \texttt{color\_hard} and \texttt{video\_easy} benchmarks of DMControl-GB.}
    \vspace{-0.2in}
    \label{fig:samples-dmc-gb}
\end{figure}

We open-source \emph{DMControl Generalization Benchmark} (DMControl-GB), a benchmark for generalization in vision-based RL, based on the DeepMind Control suite \cite{deepmindcontrolsuite2018}. In DMControl-GB, agents are trained in a fixed environment (denoted the \emph{training} environment), and the generalization ability of agents is evaluated on unseen, visually diverse test distributions. Samples from each of the proposed benchmarks are shown in Figure \ref{fig:samples-dmc-gb}. We benchmark a number of recent, state-of-the-art methods for sample-efficiency and generalization: (i) CURL \cite{srinivas2020curl}, a contrastive representation learning method; (ii) RAD \cite{laskin2020reinforcement}, a study on data augmentation for RL that uses random cropping; and (iii) PAD \cite{hansen2020deployment}, a method for self-supervised policy adaptation. All algorithms are implemented in a unified framework using standardized architecture and hyper-parameters, wherever applicable. DMControl-GB, SODA, and implementations of the baselines are open-sourced at \url{https://github.com/nicklashansen/dmcontrol-generalization-benchmark}.

\subsection{Additional results on DMControl-GB}
Our main results, Figure \ref{fig:dmc-curves-conv}, Figure \ref{fig:dmc-curves-overlay}, and Table \ref{tab:dmc-all}, consider the \texttt{color\_hard} and \texttt{video\_easy} benchmarks of DMControl-GB. Results from our experiments on two additional benchmarks from DMControl-GB are shown in Table \ref{tab:dmc-gb}. SODA outperforms all previous state-of-the-art methods in 9 out of 10 environments and achieves near-optimal generalization on the \texttt{color\_easy} benchmark. On the extremely challenging \texttt{video\_hard} benchmark, SODA shows substantial improvements over baselines, but there is still room for improvement on the majority of tasks. We conjecture that the considerable performance gap between the training environment and \texttt{video\_hard} is due to observational overfitting \cite{song2019observational}, and in our experiments we find that our baseline methods tend to rely on visual features such as shadows that may be correlated with the reward signal, but are not necessarily present at test-time.

\begin{table}
\caption{Average return of SODA and state-of-the-art methods on the \texttt{color\_easy} and \texttt{video\_hard} environments from DMControl-GB. Mean and std. deviation of 5 runs. Results for \texttt{color\_hard} and \texttt{video\_easy} are shown in Table \ref{tab:dmc-all}.}
\label{tab:dmc-gb}
\centering
\resizebox{0.425\textwidth}{!}{%
\begin{tabular}{lcccc}
\toprule
DMControl-GB    & CURL & RAD  & PAD    & SODA \\
(color\_easy)           & \cite{srinivas2020curl} & \cite{laskin2020reinforcement} & \cite{hansen2020deployment}  & (overlay) \\ \midrule
\texttt{walker,}        & $645$ & $636$ & $687$ & $\mathbf{811}$ \vspace{-0.75ex} \\
\texttt{walk}            & $\scriptstyle{\pm55}$ & $\scriptstyle{\pm33}$ & $\scriptstyle{\pm119}$ & $\mathbf{\scriptstyle{\pm41}}$ \vspace{0.75ex} \\
\texttt{walker,}        & $866$ & $807$ & $894$ & $\mathbf{960}$ \vspace{-0.75ex} \\
\texttt{stand}            & $\scriptstyle{\pm46}$ & $\scriptstyle{\pm67}$ & $\scriptstyle{\pm39}$ & $\mathbf{\scriptstyle{\pm4}}$ \vspace{0.75ex} \\
\texttt{cartpole,}        & $668$ & $763$ & $812$ & $\mathbf{859}$ \vspace{-0.75ex} \\
\texttt{swingup}            & $\scriptstyle{\pm74}$ & $\scriptstyle{\pm29}$ & $\scriptstyle{\pm20}$ & $\mathbf{\scriptstyle{\pm15}}$ \vspace{0.75ex} \\
\texttt{ball\_in\_cup,}        & $565$ & $727$ & $775$ & $\mathbf{969}$ \vspace{-0.75ex} \\
\texttt{catch}            & $\scriptstyle{\pm168}$ & $\scriptstyle{\pm87}$ & $\scriptstyle{\pm159}$ & $\mathbf{\scriptstyle{\pm3}}$ \vspace{0.75ex} \\
\texttt{finger,}        & $781$ & $789$ & $\mathbf{870}$ & $855$ \vspace{-0.75ex} \\
\texttt{spin}            & $\scriptstyle{\pm139}$ & $\scriptstyle{\pm160}$ & $\mathbf{\scriptstyle{\pm54}}$ & $\scriptstyle{\pm93}$ \vspace{0.75ex} \\\bottomrule
\toprule
DMControl-GB    & CURL & RAD  & PAD    & SODA \\
(video\_hard)           & \cite{srinivas2020curl} & \cite{laskin2020reinforcement} & \cite{hansen2020deployment}  & (overlay) \\ \midrule
\texttt{walker,}        & $58$ & $56$ & $93$ & $\mathbf{381}$ \vspace{-0.75ex} \\
\texttt{walk}            & $\scriptstyle{\pm18}$ & $\scriptstyle{\pm9}$ & $\scriptstyle{\pm29}$ & $\mathbf{\scriptstyle{\pm72}}$ \vspace{0.75ex} \\
\texttt{walker,}        & $45$ & $231$ & $278$ & $\mathbf{771}$ \vspace{-0.75ex} \\
\texttt{stand}            & $\scriptstyle{\pm5}$ & $\scriptstyle{\pm39}$ & $\scriptstyle{\pm72}$ & $\mathbf{\scriptstyle{\pm83}}$ \vspace{0.75ex} \\
\texttt{cartpole,}        & $114$ & $110$ & $123$ & $\mathbf{429}$ \vspace{-0.75ex} \\
\texttt{swingup}            & $\scriptstyle{\pm15}$ & $\scriptstyle{\pm16}$ & $\scriptstyle{\pm24}$ & $\mathbf{\scriptstyle{\pm64}}$ \vspace{0.75ex} \\
\texttt{ball\_in\_cup,}        & $115$ & $97$ & $66$ & $\mathbf{327}$ \vspace{-0.75ex} \\
\texttt{catch}            & $\scriptstyle{\pm33}$ & $\scriptstyle{\pm29}$ & $\scriptstyle{\pm61}$ & $\mathbf{\scriptstyle{\pm100}}$ \vspace{0.75ex} \\
\texttt{finger,}        & $27$ & $34$ & $56$ & $\mathbf{302}$ \vspace{-0.75ex} \\
\texttt{spin}            & $\scriptstyle{\pm21}$ & $\scriptstyle{\pm11}$ & $\scriptstyle{\pm18}$ & $\mathbf{\scriptstyle{\pm41}}$ \vspace{0.75ex} \\\bottomrule
\end{tabular}
}
\vspace{-0.05in}
\end{table}

\section{IMPLEMENTATION DETAILS}
\label{sec:appendix-hyperparams}

Table~\ref{tab:hyperparameters-dmc} summarizes the hyper-parameters of SODA and baselines used in our DMControl-GB experiments. We use identical algorithms, architectures, and hyper-parameters for robotic manipulation, but opt for observations as stacks of $k=3$ frames in DMControl-GB and $k=1$ in robotic manipulation. We implement SODA and baselines using Soft Actor-Critic (SAC) \cite{sacapps} as the base algorithm, and adopt network architecture and hyperparameters from \cite{hansen2020deployment}. We apply the same architecture and hyperparameters in all methods, wherever applicable. As the action spaces of both DMControl-GB and the robotic manipulation task are continuous, the policy learned by SAC outputs the mean and variance of a Gaussian distribution over actions. Frames are RGB images rendered at $100\times100$ and cropped to $84\times84$ as in \cite{srinivas2020curl, laskin2020reinforcement, hansen2020deployment}, such that inputs to the network are of size $3k\times84\times84$, where the first dimension indicates the color channels and the following dimensions represent spatial dimensions. The same crop is applied to all frames in a stack. The shared encoder consists of 11 convolutional layers and outputs features of size $32\times21\times21$. In our SODA implementation, both $g_{\theta}, h_{\theta}$ are implemented as MLPs with batch normalization \cite{Ioffe2015BatchNA} and we use a batch size of 256 for the SODA task. Projections are of dimension $K=100$, and the target components $f_{\psi}, g_{\psi}$ are updated with a momentum coefficient $\tau = 0.005$. $\mathcal{L}_{RL}$ and $\mathcal{L}_{SODA}$ are optimized using Adam \cite{Kingma2015AdamAM}, and in practice we find it sufficient to only make a SODA update after every second RL update, i.e. $\omega=2$.

\begin{table}
\parbox{0.47\textwidth}{
\caption{Hyper-parameters for the DMControl-GB tasks.}
\label{tab:hyperparameters-dmc}
\centering
\resizebox{0.47\textwidth}{!}{%
\begin{tabular}{@{}ll@{}}
\toprule
Hyper-parameter                                                                   & Value                                                                             \\ \midrule
Frame rendering                                                                  & $3\times100\times100$                                                                     \\
Frame after crop                                                                 & $3\times84\times84$                                                                       \\
Stacked frames                                                                   & 3                                                                                 \\
Number of conv. layers                                                           & 11                                                                           \\
Number of filters in conv.                                                       & 32
                                \\
\begin{tabular}[c]{@{}l@{}}Action repeat\\~\\~ \end{tabular}                                                                    & \begin{tabular}[c]{@{}l@{}}2 (\texttt{finger\_spin})\\ 8 (\texttt{cartpole\_swingup})\\ 4 (otherwise)\end{tabular} \\
Discount factor $\gamma$                                                         & 0.99                                                                              \\
Episode time steps                                                       & 1,000                                                                                        \\
Learning algorithm                                                            & Soft Actor-Critic                                                                                               \\
Number of training steps                                                            & 500,000                                                                           \\
Replay buffer size                                                               & 500,000                                                                           \\
Optimizer (RL/aux.)                                                & Adam ($\beta_1=0.9, \beta_2=0.999$)                                                       \\
Optimizer ($\alpha$)                                                             & Adam ($\beta_1=0.5, \beta_2=0.999$)
                                     \\
Learning rate (RL)                                                               & 1e-3
                                    \\
Learning rate ($\alpha$)                                                         & 1e-4                                                                              \\
Learning rate (SODA)                                                             & 3e-4                                                                              \\
Batch size (RL)                                                                  & 128                                                                               \\
Batch size (SODA)                                                                & 256                                                                               \\
Actor update freq.                                                              & 2                                                                                 \\
Critic update freq.                                      & 1                                                                                 \\
\begin{tabular}[c]{@{}l@{}}Auxiliary update freq. \\~\\~ \end{tabular}                                                                    & \begin{tabular}[c]{@{}l@{}}1 (CURL)\\ 2 (PAD/SODA)\\ N/A (otherwise)\end{tabular} \\
Momentum coef. $\tau$ (SODA)                                      & 0.005                                                                                 \\ \bottomrule
\end{tabular}%
}
}
\end{table}

\section{DATA AUGMENTATIONS}
\label{sec:appendix-data-augmentations}

\begin{figure}
    \centering
    \begin{subfigure}[b]{0.48\textwidth}
        \centering
        \includegraphics[width=\textwidth]{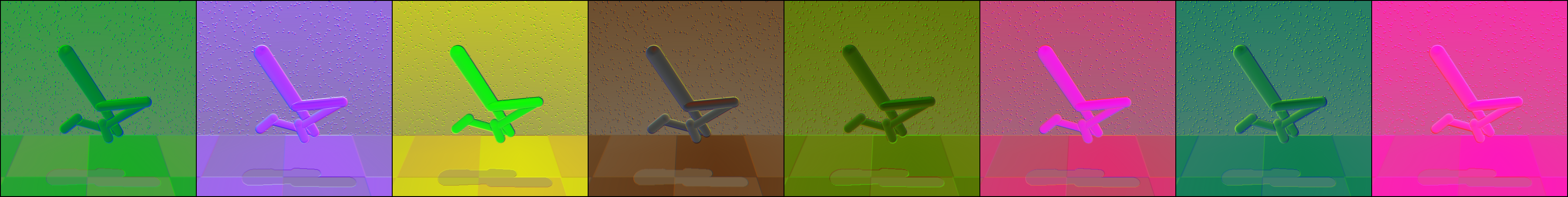}
        \caption{\emph{Random convolution} data augmentation.}
        \label{fig:samples-dmc-aug-conv}
    \end{subfigure}
    \begin{subfigure}[b]{0.48\textwidth}
        \centering
        \vspace{2ex}
        \includegraphics[width=\textwidth]{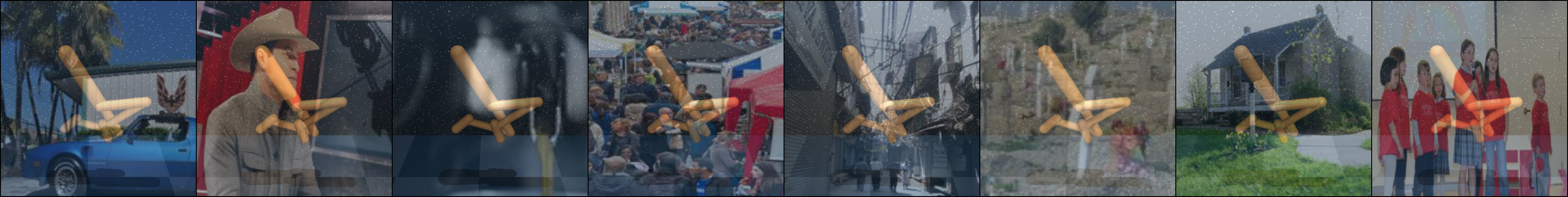}
        \caption{\emph{Random overlay} data augmentation.}
        \label{fig:samples-dmc-aug-overlay}
    \end{subfigure}%
    \caption{\textbf{Samples from data augmentation.} In this work, we consider the \emph{random convolution} data augmentation shown in (a), as well as the novel \emph{random overlay} shown in (b). }
    \vspace{-0.2in}
    \label{fig:samples-dmc-aug-conv-overlay}
\end{figure}

Section \ref{sec:data-augmentation} discusses two data augmentations that we consider in this work: (1) \emph{random convolution} \cite{Lee2019ASR, laskin2020reinforcement}, where a randomly initialized convolutional layer is applied; and (2) \emph{random overlay}, where we linearly interpolate between an observation $o$ and an image $\varepsilon$ to produce an augmented view $o' = t_{overlay}(o)$, as formalized in Equation \ref{eq:random-overlay}. In practice, we use an interpolation coefficient of $\alpha=0.5$ and sample images from Places \cite{zhou2017places}, a dataset containing 1.8M diverse scenes. Samples from the two data augmentations are shown in Figure \ref{fig:samples-dmc-aug-conv-overlay}. 

\end{document}